\documentclass[sigconf]{acmart}

\usepackage{booktabs} 
\usepackage{graphics}
\usepackage{amsmath}
\usepackage{amsfonts}
\usepackage{epsfig}
\usepackage{url}
\usepackage{multirow}
\usepackage{graphicx}
\usepackage{xspace}
\usepackage{wasysym,algorithm}
\usepackage[noend]{algpseudocode}
\usepackage{enumerate}
\usepackage{enumitem}
\usepackage{algcompatible}
\usepackage[bottom]{footmisc}
\usepackage{color}
\usepackage{pdfpages}

\newcommand{\alg}{\textsc{Guided-EpiDeep}\xspace}
\newcommand{\prob}{\textsc{Guided Epidemic Forecasting}\xspace} 

\newcommand{\zs}{Z_{\text{smooth}}}
\newcommand{\zr}{Z_{\text{regional}}}
\newcommand{\f}{f}
\newcommand{\smoothCase}{Holiday correctness}
\newcommand{\regionalCase}{Regional equity}

\allowdisplaybreaks  

\setcopyright{none}
\copyrightyear{2020}
\acmYear{2020}

\acmConference[epiDAMIK 2020]{3rd epiDAMIK ACM SIGKDD International Workshop on Epidemiology meets Data Mining and Knowledge Discovery}{Aug 24, 2020}{San Diego, CA}
\acmBooktitle{epiDAMIK 2020: 3rd epiDAMIK ACM SIGKDD International Workshop on Epidemiology meets Data Mining and Knowledge Discovery}
\acmPrice{15.00}

\begin{document}
\title{Incorporating Expert Guidance in Epidemic Forecasting}  


\author{Alexander Rodr\'iguez$^*$,  Bijaya Adhikari$^\dagger$, Naren Ramakrishnan$^+$ and B. Aditya Prakash$^*$}
\affiliation{$^*$College of Computing, Georgia Institute of Technology}
\affiliation{$^\dagger$Department of Computer Science, University of Iowa}
\affiliation{$^+$Department of Computer Science, Virginia Tech}
\affiliation{Email: \{arodriguezc, badityap\}@cc.gatech.edu, bijaya-adhikari@uiowa.edu, naren@cs.vt.edu}

\begin{abstract}
Forecasting influenza like illnesses (ILI) has rapidly progressed in recent years from an art to a science with a plethora of data-driven methods. While these methods have achieved qualified success, their applicability is limited due to their inability to incorporate expert feedback and guidance systematically into the forecasting framework. We propose a new approach leveraging the Seldonian optimization framework from AI safety and demonstrate how it can be adapted to epidemic forecasting.
We study two types of guidance: smoothness and regional consistency of errors, where we show that by its successful incorporation, we are able to not only bound the probability of undesirable behavior to happen, but also to reduce RMSE on test data by up to 17\%.
\end{abstract}

%
%



\maketitle

\section{Introduction}
\label{sec:intro}

Epidemic outbreaks incur heavy burden in terms of both health and economic costs (like the ongoing 2019-Covid corona virus epidemic). According to the world health organization (WHO), more than 15 thousand lives were lost due to the Ebola outbreak in West Africa between 2013 and 2016\footnote{http://apps.who.int/gho/data/view.ebola-sitrep.ebola-summary-latest}\!. The economic cost of Ebola is estimated to be more than 53 billion dollars\footnote{https://www.reuters.com/}\!. Timely forecasting of epidemic outbreaks is critical. 
Accurately forecasting various metrics of an epidemic outbreak informs practitioners and policymakers about impending scenarios and helps them devise strategic countermeasures, such as quarantining subpopulations, increasing vaccination availability, and school closures.

\begin{figure}[t]
    \centering
  \begin{tabular}{cc}
    \includegraphics[width=0.25\textwidth]{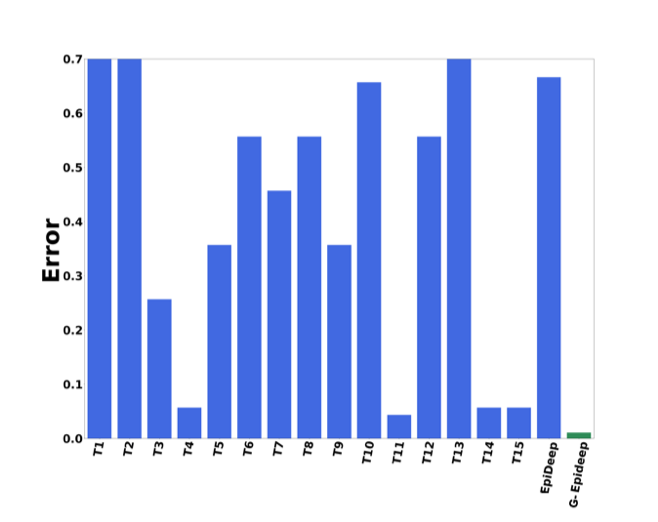} &
    \includegraphics[width=0.25\textwidth]{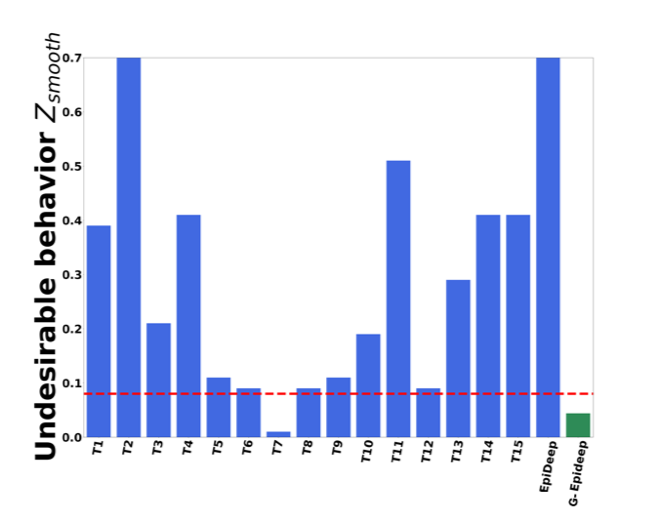} \\
    (a)  Error  & (b) Guidance \\
  \end{tabular}
  
    \caption{Comparison of approaches in terms of (a) error in forecasting and (b) a guidance metric. In both plots lower is better. The red line in (b) is the threshold determined by guidance. T1 to T14 is the performance of teams participating in the 2015 FluSight challenge. Our method Guided-Epideep (GEpideep in the plot) is the only method which satisfies the guidance and gives the  lowest prediction performance error.}
    \label{fig:epsilon_intro}
\end{figure}

In this paper, we focus on
influenza
forecasting, motivated by the CDC FluSight prediction challenge~\cite{biggerstaff2016results} which seeks to predict the inclidence of Influenza-like-Illnesses (ILI) in the US. Influenza is a major disease in the United States and beyond, causing thousands of fatalities every year. ILI is a symptomatic definition of illnesses that serves as a bellwether for real
influenza incidence in a population. There has been a surge in recent research interest in influenza forecasting giving rise to a variety of  mechanistic~\cite{zhang2017forecasting,shaman2010absolute} and statistical approaches~\cite{adhikari2019epideep,brooks2015flexible}. 
Mechanistic approaches predict influenza burden using simulation and aggregation of large epidemiological models. These  models require a lot of calibration and hence are limited by their parameters to generalize well and fit the data~\cite{nsoesie2014systematic}. Hence many researchers have begun exploring statistical approaches for this task, which train on historical ILI data and use the trained model to make forecasts for the current season. 

Influenza seasons tend to be highly dynamic and have high variability due to numerous factors (e.g., weather, human mobility, virus strains circulating amongst the population) affecting the overall characteristics of the season. Moreover, different seasons and regions have different dominating influenza virus types. Further, the surveillance data collected (using ILINet) is a composite of multiple sources, is non-uniform, and is biased in many domain-specific ways.
Hence while statistical approaches can frequently perform more accurate predictions than mechanistic models, they often show undesirable, unexplainable, or otherwise unexpected behavior. 

For example, consider influenza incidence during the annual holiday season in the US. During
this period, patients typically self-select and refrain from going to health providers, unless the situation is serious. This causes a temporary drop in recorded ILI incidence. However, as human mobility is high, flu activity rapidly  increases in the following weeks. This can not be modeled using standard mechanistic epidemiological models~\cite{osthus2019dynamic}. At the same time, 
statistical approaches `over-correct' and exaggerate the temporary `dip'. 
Hence if we can ensure that the forecasting model's predictions are reasonably 'smooth', such a behavior can be avoided. This 'smoothness' of the forecasts is well-motivated from other epidemiological considerations as well. 
As Figure \ref{fig:epsilon_intro} (b) shows, almost all the  methods used in the 2015 CDC FluSight challenge show this 'lack of smoothness' behavior (lower is better). 

To tackle such issues, in this paper, we propose incorporating \textit{expert guidance} into statistical models for epidemic forecasting. 
In the case above, may be the expert can give the guidance that week-to-week forecast should be smooth, which can alleviate the over-correction problem.
Indeed, incorporating this guidance helps our approach outperform the baselines while maintaining accuracy. Our approach 'Guided EpiDeep' is the only method to show desirable behavior (having guidance metric $Z_{smooth}$ below the predefined threshold (red line)) while also getting the lowest errors (Figure \ref{fig:epsilon_intro}). 

The Christmas effect described above is just one example of how an epidemiological expert's domain knowledge can be brought to bear upon influenza forecasting. 
If such expert insights can be incorporated in the forecasting framework, the  resulting forecasts will be better, actionable, and insightful. As another example, an expert will also able to guide the forecasting model to demonstrate desirable behaviors, as undesirable prediction could lead to undesirable health policies.
For example, if rural regions have systematically lower prediction accuracy in forecasting models than other, more urban regions, that will lead to systematic under-allocation of resources for certain regions. 
Incorporating expert guidance to prevent this will allow the framework to have behavior that is more desirable from a policy viewpoint as well.

To design a forecasting framework as envisioned here, there are several challenges.
The first challenge is (a) how to design a general framework for any influenza statistical forecasting model to ingest and leverage expert guidance. Designing a general framework to incorporate guidance allows existing approaches to include expert guidance. 
The second challenge is (b) how to ensure that the framework is easy to use and generates useful feedback to the user. Moreover, the framework should communicate the extent to which the guidance was successfully incorporated and whether the guidance is helpful or not. Such a framework will aid in selecting guidance and make the forecast interpretable with respect to the guidance provided. None of the existing approaches is able to tackle these challenges. 

In this paper, we leverage the Seldonian Optimization framework proposed in AI safety to enforce expert guidance (desired behavior) and prevent undesirable behavior. 
Our framework provides feedback to the user regarding the success or failure in the incorporating the expert insights. In case the framework fails to incorporate the insight, it communicates the failure to the user, who in turn can take steps to alter/improve the insight or change data or modify model hyper-parameters. Our contributions are as follows.
\begin{itemize}
    \item \textbf{Novel method for incorporating expert guidance: } We explore a novel problem \& adapt a successful framework to obtain domain-based consistency (and guidance), and perform extensive experiments to show properties of the framework.
    \item \textbf{Flexible user interaction framework:} The framework adapts to the user's requirements.
    \item \textbf{Real data case-study: }We present concrete case studies showing examples of expert guidance motivated by epidemiologist observation, and how our method helps to achieve experts requirements.
\end{itemize}

The rest of the paper is organized in the following way: we first motivate our problem, and formulate it. Then we present our method, and then empirical studies on real CDC data. We finally end with related work and conclusions.

\section{Problem Formulation}
\label{sec:prob}
In this section, we introduce the novel problem of aiding statistical epidemic forecasting models with an expert's guidance. Before, we formalize our problem, we present the problem setting.

\subsection{Epidemic Forecasting}
Motivated by the setup of the CDC FluSight challenge, we study the epidemic forecasting problem from a temporal seasonality standpoint, such as in influenza.
For this problem, we are given data $\mathcal{D}$ in the form of time series (e.g. the wILI burden per week for every season) and a predictive task $\mathcal{T}_w$, which sets what the target is and the time $w$ (usually a week) when this prediction is to be made.
Examples of targets are immediate-future incidences, peak intensity for season $i$, and the time when the peak value occurs. 

The annual FluSight Challenge hosted by CDC asks to forecast metrics related to the current influenza season for the national and regional levels~\cite{biggerstaff2016results,biggerstaff2018results}. The CDC releases influenza surveillance data, referred to as weighted Influenza-like Illness (wILI), each week for every region. Given the latest partially observed influenza season, often represented as a time-series, the challenge asks to perform four different types of prediction tasks $\mathcal{T}_w$. They involve forecasting the incidence (wILI) value for the next four weeks, the onset of the season, the peak incidence value, and the time when the peak occurs. wILI incidence curves for each season since 1997/98 are publicly available\footnote{https://gis.cdc.gov/grasp/fluview/fluportaldashboard.html}.

\subsection{Expert guidance}
Expert guidance for epidemic forecasting is about leveraging multiple forms of domain knowledge and other preferences. An expert may want to guide a statistical model based on many considerations. Such considerations may include the epidemiology of the disease, characteristics of active virus strains (e.g. transmissability, reproduction rate), activity intensity in other other latitudes that dealt with the same virus strain, or efficacy of the vaccines to active strains of virus. It can also include some auxiliary knowledge. For example, it is well known that the Christmas holiday season in the US has specific impact on the flu spread which can not be captured by regular mechanistic epidemiological models~\cite{osthus2019dynamic}. 
It can include other public health policy considerations too, to ensure desirable behaviors like fairness in resource allocations. We give two specific examples we will use in our paper.

\par\noindent\textbf{Example 1: \smoothCase.} 
As mentioned earlier, during the holiday season recorded epidemic activity temporarily drops due to patients' tendency to not seek healthcare. However, an expert notices that predictions of current statistical models are not 'smooth' i.e. they change a lot week-week and 'over-correct' during this time (a fact we demonstrate later in our experiments using the predictions of all the teams which participated in the CDC FluSight 2015 challenge). Hence s/he may want to more accurately forecast influenza incidence during the Holiday season incorporating the `smoothness' property. 

\par\noindent\textbf{Example 2: \regionalCase.}
In this case, a CDC expert wants to make sure that forecasting errors across rural/urban regions are not lopsided and are balanced, so that intervention resources can be distributed more fairly. We again demonstrate that existing models (from the CDC FluSight challenge) fail to have this important property. Hence the expert wants to ensure that a statistical forecasting model has this property, but also maintain accuracy.

\subsection{Desired Properties of Guidance}
In this paper, we focus on incorporating such types of guidance into statistical epidemic forecasting models. Designing a framework which can incorporate guidance must be able to exhibit some ideal properties, especially as it is meant for experts who may not know the internals or any technical details of the statistical models. 
\begin{itemize}

  \item  \textbf{P1.} Promote one or more desirable behaviors during training. 
  \item  \textbf{P2}. Have a mechanism to guarantee tolerance on deployment.
  \item \textbf{P3.} Be flexible to any generic ad-hoc guidance and be compatible with state of the art epidemic forecasting models.
  \item  \textbf{P4.} Be easy to use for the user/expert.
  \item  \textbf{P5.} Provide feedback to user if guidance could not be incorporated.
  
\end{itemize}

Let us discuss the underlying reasons behind the importance of each of the properties described above.
P1 is an important facet of guidance: when training, the preference should be given to candidate models aligned with guidance goals. In addition, an ideal framework should be able to enforce more that one desirable behavior. Once the training completed, one can not expect that guidance will be met in unseen data at all times. Hence, it is natural to think about a probability of the trained model in meeting the guidance in unseen data. 
P2 sets guarantees on the expected probability of a model to exhibit the desirable behavior on unseen (test) data. 
P3 takes into consideration that experts' requirements may be related to any characteristic of the epidemic season. Furthermore, to leverage existing statistical forecasting models, the framework should provide a path to easily incorporate them. 
P4 aims to provide the user an easy interface to leverage the framework, treating the statistical model essentially as a black box. 
Finally, P5 importantly aspires to clearly communicate the result of attempting to incorporate the proposed guidance.
Note that in our context, expert guidance can be motivated by practical considerations, and the data is really a composite signal (ILI cases rather than exactly flu cases). Hence sometimes expert guidance can indeed not be borne out by data, or be 'completely wrong' (unlike theory-guided data science \cite{karpatne_theory-guided_2017}, where scientific knowledge is considered ground truth) -- so our framework needs a principled mechanism to signal this fact and provide feedback. The feedback provided opens possibilities to fruitful interactions as the expert may explore with different behaviors and tolerances to find the most suitable, and even test 'what-if' scenarios.

\subsection{Definitions} Taking these properties in consideration, we make the following definitions to then state our problem.

\begin{definition}
\textit{Expert guidance:} 
We represent expert guidance as a tuple $\langle g, \delta \rangle$, where $g: \Theta \rightarrow  \mathbb{R}$ is a function that maps a candidate forecasting model $\theta \in \Theta$ to a measure of desirable or undesirable behavior of $\theta$, 
and $\delta \in [ 0, 1]$ is a tolerance which constraints the probability of the model to exhibit this behavior.

\label{def:guidance}
\end{definition}
\begin{definition}
\textit{Successful incorporation of guidance:} 
We successfully incorporate guidance when we obtain a forecasting model $\theta$ for which our desired tolerance is met. 
\label{def:sucesss}
\end{definition}

Note that our definition of expert guidance allows any framework which adopts it to exhibit all five desirable properties. Since the function $g$ encodes one or more desirable behavior quantitatively, it can be used to enforce the behavior, satisfying P1. The parameter $\delta$ is the tolerance of undesirable behavior as mentioned in P2. Our definition of guidance is general enough to incorporate wide range of user insights to meet property P3. The only requirement is that the deviation from the desired behavior needs to be captured by the function $g$. Similarly, the user/expert do not need to be aware of underlying optimization framework and statistical model to incorporate the guidance as the function $g$ is independent of both, satisfying P4. Similarly, if the value of the function $g$ is greater than the threshold $\delta$, the framework can communicate with the user regarding its inability to meet the guidance. We show how we can adapt our examples given before using our framework later (in Section~\ref{sec:cbc}).

\subsection{Problem Statement}

Having defined the notion of guidance that meets all the desired properties, we can state our problem as follows:

\prob{}: \textit{
Given a forecasting model which defines hypothesis space $\Theta$, 
data $\mathcal{D}$, 
a predictive task $\mathcal{T}_w$, 
and expert guidance $\langle  g(\theta), \delta \rangle$, 
we are required to 
return an optimal model $\theta$, if found, that successfully incorporates expert guidance or return feedback that such $\theta$ could not be found.
}
 
In this paper, the predictive task we consider is the future incidence forecasting. Our task $\mathcal{T}_w$ asks for influenza incidence at week $w+1$ given that the incidence till week $w$ is observed. And as the problem states, our goal is to enforce expert guidance, while solving for the predictive task. However our framework can easily handle other predictive tasks as well (like peak prediction etc).

\section{Our Method}
\label{sec:methods}

As stated above, the \prob{} problem requires a base forecasting model upon which the guidance is enforced. To enforce the guidance, we need a framework which optimizes for performance with respect to the predictive task $\mathcal{T}_w$ as well as ensures that the constraint imposed by the guidance $\langle  g(\theta), \delta \rangle$ is met. Here we leverage Seldonian Optimization which does this. 

\subsection{Seldonian Optimization}
\label{subsec:sop}
The Seldonian optimization framework~\cite{thomas2019preventing} was recently proposed for Artificial Intelligence (AI) safety. It is designed to prevent AI models from showing undesirable behavior such as gender or racial bias. Traditional AI algorithms optimize an objective function to select a model $\theta$ as a solution from the space of all possible models $\Theta$. This framework precludes undesirable behavior of AI model by enforcing behavioral constraints on the optimization objective. Hence, a probabilistic constraint is added to the optimization such that the probability that the value of a predefined undesirable behavior metric $g(\theta)$ is greater than $0$. 
After training, to ensure the behavioral constraint will be met when the solution is deployed, this framework has a safety test mechanism, which is performed in unseen data.
If the model meets the requirements of the safety test, the trained model is returned, else the framework returns no solution found (NSF).

A natural question that arises is what kind of base forecasting model (which is required by our problem) best works with the Seldonian optimization framework. Intuitively, models which learn/train by back-propagating errors are most suited for the Seldonian Framework as it learns through back-propagating as well. Hence, here we chose a recently proposed deep learning based influenza forecasting model EpiDeep~\cite{adhikari2019epideep} as the base model upon which the guidance is to be enforced. We desribe Epideep briefly next. However, we wish to emphasize that our framework is general.

\subsection{EpiDeep}
Epideep~\cite{adhikari2019epideep} is a recent deep neural architecture designed specifically for influenza forecasting. It exploits  seasonal similarity between the current season and historical seasons via deep clustering~\cite{xie2016unsupervised}. 
The clustering module learns a latent low dimensional embedding of the seasons, such that the similarity between the seasons in the embedding space is meaningful for the task at hand. The clustering module in EpiDeep is designed such that it is possible to learn the embedding of the partially observed current season in the space of fully observed historical seasons. Epideep also uses long short-term memory (LSTM)~\cite{gers1999learning} to encode in-season patterns of the current season. It then combines the embeddings from the clustering module and the LSTM and feeds the aggregated embedding to the decoder module, which make predictions for task $\mathcal{T}_w$.
For the set of seasons $\mathcal{S}$ where each season $S \in \mathcal{S}$ is represented as a time series $S = s_1, s_2, \dots, s_T$ in the training season, to predict the incidence observed in week $w$  EpiDeep is trained with a loss function $\mathcal{L}(\theta) = \sum_{S \in \mathcal{S}} ||\hat{y} - s_w||+ \beta$ , where $\theta \in \Theta$ is the trained model,$\hat{y}$ is the prediction made by $\theta$ and $s_w$ is the observed incidence and $\beta$ is the internal loss for Epideep not directly related to the task $T_w$. Note that while training only the weeks prior to week $w$ is leveraged.

\subsection{Expert-guided EpiDeep}

The next natural question is how to adapt the Seldonian optimization framework to train EpiDeep with expert guidance. Before we answer that, let us define some notations.
Let us have several different expert guidance to incorporate $\{ \langle g_i, \delta_i \rangle \}_{i=1}^{n} $. 
We adopt the convention that if $g_i(\theta)\leq \epsilon$ for some small $0 \leq \epsilon$, the forecasting model $\theta$ does not exhibit undesirable behavior. Hence we impose probabilistic constraint on $g_i(\theta)$, on the model optimization.  Hence, our updated optimization objective is as follows.
\begin{eqnarray} 
 \arg\min_{\theta} \sum_{S \in \mathcal{S}} ||\hat{y} - s_w||+ \beta \nonumber\\ 
s.t.  \Pr\left(g_{i}(\theta) \leq \epsilon \right) \geq 1-\delta_{i}, \forall i \in \{ 1, \ldots, n \}
\label{eqn:objective}
\end{eqnarray} 

Here, $\hat{y}$ is the prediction made by model $\theta$ for the prediction task $\mathcal{T}_w$. The objective above indicates that we want to ensure the probability that the desirable behavior (i.e, $g_i(\theta)\leq\epsilon$) occurs is greater than $1-\delta_{i}$ for some small $0 \leq \delta \leq 1$, while the difference between predicted incidence value and the eventually observed value is minimized. Note that, we also want to ensure that the probability that the desirable behavior holds even in test/deployment stage.

Following~\cite{thomas2019preventing}, we ensure that our approach optimizes the objective while not violating the constraints and it generalizes to other unseen data with high confidence. It does so by dividing the given training data $\mathcal{D}$ into two partitions $D_c$ and $D_s$. $D_c$ partition is used for the model selection/optimization, while $D_s$ partition is only used to verify that the guidance behavior is met in unobserved data. If the guidance behavior is not met in $D_s$, then the framework ensures that no model is returned. In Algorithm~\ref{alg:seldonian}, we leverage the Seldonian framework to design our algorithm Guided EpiDeep as follows.
{
\begin{algorithm}[ht!]{}
\caption{Guided EpiDeep}
\begin{algorithmic}[1]
\small
    \STATE \textbf{Input:} $\mathcal{D}$, $ \langle g, \delta \rangle $, $U_{\mathcal{L}}$.
    \STATE Partition $\mathcal{D}$ into $D_{c}$ (for candidate selection) and $D_{s}$ (for safety test)
    \STATE $\left.\theta_{c} \in \arg \min _{\theta \in \Theta} \text{CandidateLossFunction}(D_{c}, \delta, \epsilon, U_{\mathcal{L}},\left|D_{s}\right|\right)$
     \STATE \{ Safety test using $D_s$ \}
    \IF{UpperBound($\left.\theta_{c}, D_{s}, \delta, U_{\mathcal{L}}\right) \leq \epsilon$}
        \STATE return $\theta_{c}$
    \ELSE
      \STATE return No Solution Found (NSF)
    \ENDIF
\end{algorithmic}
\label{alg:seldonian}
\end{algorithm}
}

Here, the function UpperBound in line 5 measures if the behavior of candidate model $\theta_c$ is desirable as per the guidance provided. Based on predictions made by $\theta_c$, variable $Z$ is defined to quantify the deviation from the desirable behavior for each prediction made. We discuss how variable $Z$ is constructed for the guidance we consider in Section \ref{sec:cbc}. Once $Z$ is defined, we assume it follows a normal distribution and compute UpperBound as suggested in~\cite{thomas2019preventing}. We employ an empirical upper bound on the magnitude of $\mathcal{L}$, which is denoted as $U_{\mathcal{L}}$. This bound is necessary to prevent gradient explosion when switching losses in our CandidateLossFunction.
Next we present the CandidateLossFunction subroutine in line 3 of Algorithm \ref{alg:candidate}.

In the CandidateLossFunction subroutine, we use the $D_c$ partition of the training data to train on both the objective with respect to the task $\mathcal{T}_w$ and to ensure that the returned model, $\theta$ is consistent with the guidance. To do so, variable $Z = \{Z_i| \forall i \in D_c \}$ is created using the predictions made by the model $\theta$. Then the upper bound on variable $Z$ is computed as in~\cite{thomas2019preventing}. If the upper bound computed is less than $\epsilon$, indicating that the model is showing desirable behavior with respect to the guidance, then loss on $\mathcal{L}(\theta)$ is returned else, the loss on the bound is returned. Note that internally, $\lambda \in R_{>=0}$ balances the trade-off between loss and guidance.
{
\begin{algorithm}[ht!]{}
\caption{CandidateLossFunction}
\begin{algorithmic}[1]
\small
    \STATE \textbf{Input:} Candidate $\theta_c$, $D_c$, $ \langle g, \delta \rangle $, $U_{\mathcal{L}}$, $\left|D_{s}\right|$
    \STATE Create an array of $Z_i$, where $i \in D_c$ 
     \STATE $\hat{U}=\text{PredictedBound}(Z_i, \delta , \left|D_{s}\right|)$
    \IF{ $\hat{U} \leq \epsilon_i$}
        \STATE return $\sum_{S \in \mathcal{S}} ||\hat{y} - s_w||  + \beta + \lambda \frac{1}{|Z|} \sum_{i=1}^{|Z|} \left|Z_i\right|$  
    \ENDIF
    \STATE return $U_{\mathcal{L}} + \hat{U} + (\lambda-1)\epsilon$ 
\end{algorithmic}
\label{alg:candidate}
\end{algorithm}
}

Now, the question is how to define the variables $Z$ for a given guidance. We discuss it next.

\subsection{Constructing Behavioral Constraints}
\label{sec:cbc}
In this paper, we select smoothness as our expert guidance for seasonal influenza forecasting. In this section, we show the construction of constraint objectives for these expert guidance in the form of the $Z$ variables. 

\subsubsection{Smoothness}
Mechanistic epidemiological models reveal that the epidemiological curves tend to be smooth with a single peak~\cite{shaman2012forecasting}. Hence, we expect epidemic influenza seasons to be generally smooth. In fact, we observe influenza incidence curve to be smooth with the consecutive values not changing drastically. Usually, incidence are low in the beginning of the season, they gradually rise till the peak is observed and then decline near monotonically. Hence, forecasting that the influenza incidence while ensuring that the predictions are smooth is a desirable property. 

The smoothness also helps in correcting the drop observed in the influenza activity during the holiday season (discusses earlier), which arises due to the artifact of data collection. The existing approaches tend to overcompensate for the drop. Enforcing smoothness in forecasts prevents such undesirable behavior. Here we describe smoothness as follows:

\begin{definition}
\textit{Smoothness} is the max allowed difference $\epsilon$ between the predicted value and its predecessor. 
\end{definition}

We have a smoothness parameter $\epsilon$, which is the maximum change allowed between current influenza incidence and the next incidence.
In simple words, we want to ensure that the probability of smoothness function being greater than $\epsilon$.
\begin{equation}
    g(\theta) = E(|\hat{y}_{t+1} - Y_t  |) \leq \epsilon 
\end{equation}

Here, $E(|\hat{y}_{t+1} - Y_t  |)$ is the expected absolute difference between the predicted incidence $\hat{y}_{t+1}$ by the model $\theta$ and the last observed incidence $Y_t$. The guidance function $g(\theta)$ quantifies the smoothness by computing the difference between predicted and the previously observed value. The equation above, highlights that the expected difference between the forecasted value and the previously observed value should be less than some constant $\epsilon$ ensuring that the forecast maintains week-to-week smoothness. Following this, we define the varaible $Z$ for smoothness as follows:
\begin{equation}
    \zs = | \hat{y}_{t+1} - Y_t  |
\end{equation}

\subsubsection{Regional Equity} The performance of forecasting model for different regions is governed by the quality of surveillance mechanisms in place in a given region. Hence forecasting models tend to perform poorer in the regions where the surveillance data is poor. Moreover, the bias in the quality of forecast due to quality of data could be magnified, leading to less accurate forecasts in regions which already have limited surveillance resources. This could result in under-preparation, which could be catastrophic or over-preparation, which is wasteful in such regions. Hence, to ensure that the regions with poor surveillance system do not suffer more financial/health burden regional equity in the quality of forecast is required. Here we define Regional Equity as follows:

\begin{definition}
\textit{Regional Equity} is the maximum allowed difference $\epsilon$ between the quality of forecast $\mu$ between any two regions. 
\end{definition}

Here the function $\mu$ could be any measure of forecast quality such as root mean squared error (RMSE), mean average prediction error (MAPE), and so on. In this paper, we use RMSE.
Now, the guidance function for smoothness with respect to quality metric $\mu$ and two arbitrary regions $R_1$ and $R_2$ can be written as follows:

\begin{equation}
    g(\theta) = E(|\mu(\theta,t+1, R_1) -\mu(\theta,t+1, R_2)  |) - \epsilon \leq 0
\end{equation}

Here,$ E(|\mu(\theta,t+1, R_1) -\mu(\theta,t+1, R_2)$ is the expected difference in quality of forecast in regions $R_1$ and $R_2$. Now, we have:

\begin{equation}
    E(|\mu(\theta,t+1, R_1) - \mu(\theta,t+1, R_2)  |) \leq \epsilon
\end{equation}

The equation above represents our desire that under the regional equity guidance, we wish to ensure that the error in forecasting for two regions $R_1$ and $R_2$ should be less than some small constant $\epsilon$. Hence, we define the variable $Z$ for regional equity as follows:

\begin{equation}
    \zr = | \mu(\theta,t+1, R_1) - \mu(\theta,t+1, R_2)  |
\end{equation}

\subsection{Expert Interaction}
\label{subsec:interaction}

As mentioned in Section \ref{sec:prob}, two of the desirable properties of a framework to incorporate guidance is that it should be easy to use (P4) and should be able to provide feedback to user (P5). 
In this section, we present how our framework can be leveraged for exploration as well as demonstrate how an user might be able to interact with the framework.

From a user perspective, our framework provides three knobs: data, model, tolerance. An user is able to decide on how to partition the data into $D_c$ for candidate model selection and $D_s$ for safety test. Similarly, the user can decide on the base model suitable for the task at hand. The final knob corresponds to the tolerance with which the failure to incorporate the provided guidance is allowed. An expert/user can interact with the system by varying the values corresponding to the knobs.

Since our model has a mechanism for the safety test, it may return 'No Solution Found' (NSF) indicating that the guidance provided could not be met given the values of the knobs. If the model returns NSF, it is an indication for the user to either consider the guidance provided or to vary the knobs. For example, if guidance related to smoothness is decided to be changed, this can be changed from $\epsilon=0.5$ to $\epsilon=1$. If tolerance is changed, confidence in guidance is changed. Hence, the model might be able to incorporate the guidance with a lower confidence on its generalizability. On the other hand, an expert can also change data by deciding to exclude some historical seasons that are preventing the guidance provided from being.

\begin{figure}[t]
    \centering
    \includegraphics[width=0.5\textwidth]{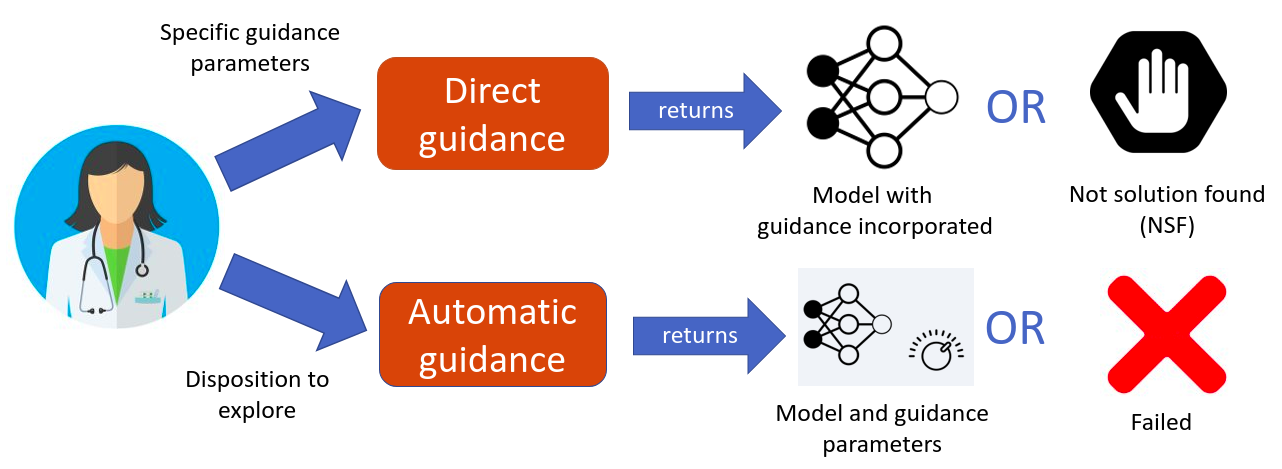}
    \caption{Flow diagram of expert interaction with Guided EpiDeep. Expert is given two modes: direct guidance and automatic guidance. The choice depends on the underlying motivation of the expert. Depending on the mode selected, the feedback is adapted to report success or failure.}
    \label{fig:diagram}
\end{figure}

For ease of usage and interaction, our framework provides two modes of usages, namely Direct guidance and Automatic guidance, and depicted in Figure~\ref{fig:diagram}. We discuss each of the usages next.

\subsubsection{Direct Guidance}
In this mode, the user specifies guidance along with all the \emph{all} parameters. Then our framework tries to incorporate the guidance within the constraints imposed by the parameters. If the framework fails to find a forecasting model which guarantees guidance incorporation, the the framework returns NSF. The direct guidance framework is presented in Algorithm~\ref{alg:seldonian}.

\subsubsection{Automatic Guidance}
The user or epidemiological expert may not have data science/mathematical background to estimate the parameters with which the guidance can be incorporated and is willing to explore.  Hence, in such cases, the framework tries to find the parameters which ensures that the guidance is met and the performance is maintained. 

Our framework is able to provide such a exploration mode for users. Here the user may specify a \emph{subset} of the parameters, and requirements in terms of performance. Our framework then explores the parameter space to find such a model. If none of the parameters explored is able to induce a model which satisfies the user requirements, the the framework returns NSF, indicating that the guidance could not be incorporated. In this paper, as an example of automatic guidance, we ask our framework to explore the parameter $\epsilon_i$ such that no compromise is made in terms of RMSE.

\section{Experiments} 
\label{sec:exp}

\subsection{Setup}
\label{subsec:setup}
We describe the experimental setup next. All experiments are conducted using a 4 Xeon E7-4850 CPU with 512GB of 1066 Mhz main memory. Our method is very fast, training for one prediction task (on 1 week) in about 3 mins. We will release the code for academic purposes.
\par\noindent\textbf{Data}
Here we use the weighted Influenza-like Illness (wILI) data released and updated by the CDC.
CDC collects the wILI data through the Outpatient Influenza-like Illness Surveillance Network (ILINet) which consists of more than 3,500 outpatient healthcare providers all over the United States.CDC defines Influenza-like Illness (ILI) as “fever (temperature of 100◦F [37.8◦C] or greater) and a cough and/or a sore throat without a known cause other than influenza.
 Weekly wILI incidence curves for each season since 1997/98 are publicly available\footnote{https://gis.cdc.gov/grasp/fluview/fluportaldashboard.html}.

\par\noindent\textbf{Research questions to address.} 
In our experiments we want to compare the performance of our approach \alg{} with the baselines for both the smoothness and regional equity guidance. We also want to evaluate the automatic guidance mode of \alg{}. Specifically, we are interested in answering the following questions.
\par\noindent Q1. Is \alg{} successful in incorporating guidance?
\par\noindent Q2. Does \alg{} give feedback?
\par\noindent Q3. Is \alg{} successful in realistic scenarios on real WILI data?

\par\noindent\textbf{Evaluation.}
Here, we use the test data $T$, which is separated out during training to evaluate the performance of \alg{}. Note that the test data is not used in either partition of the training data, namely $D_c$ used for candidate model selection and $D_s$ used for safety test.

To evaluate \alg{} with respect to Q1, we will test if the guidance is incorporated in the test data. For Q1, we train the model on $D_c$ and $D_s$ to incorporate the guidance. Once the model is trained we evaluate whether the behavior of the model in forecasting influenza season in $T$ is desirable with respect to the given guidance. To evaluate the degree to which the behaviour mandated by guidance is met in the test, we compute the probability that the behavior defined by the guidance $g(\theta)$, as defined in Section \ref{sec:methods}, falls outside the bounds. We name this metric as the failure rate of the model $\theta$. Formally, we define the failure rate as $\Pr\left( g_{i}(\theta) \right)$. To evaluate \alg{} with respect to Q2 and Q3, we perform several case-studies.

\par\noindent\textbf{Baselines.} 
We use EpiDeep for performance and state of art baselines from the FluSight challenge for our case studies to show how they perform in a real-world scenario as posed by the CDC FluSight challenge. The complete list of teams we use is presented in the supplementary.

\subsection{Direct Guidance}
\label{subsec:direct}

As mentioned earlier, in the direct guidance mode, the user/expert provides guidance as well as other parameters. Here, \alg{} searches for the model which in able to incorporate the guidance within the constraints imposed by the parameters. For direct guidance, we evaluate \alg{} in terms of Q1 and Q3.

\subsubsection{Performance}
\begin{figure*}[h]
    \centering
    \includegraphics[width=0.9\textwidth]{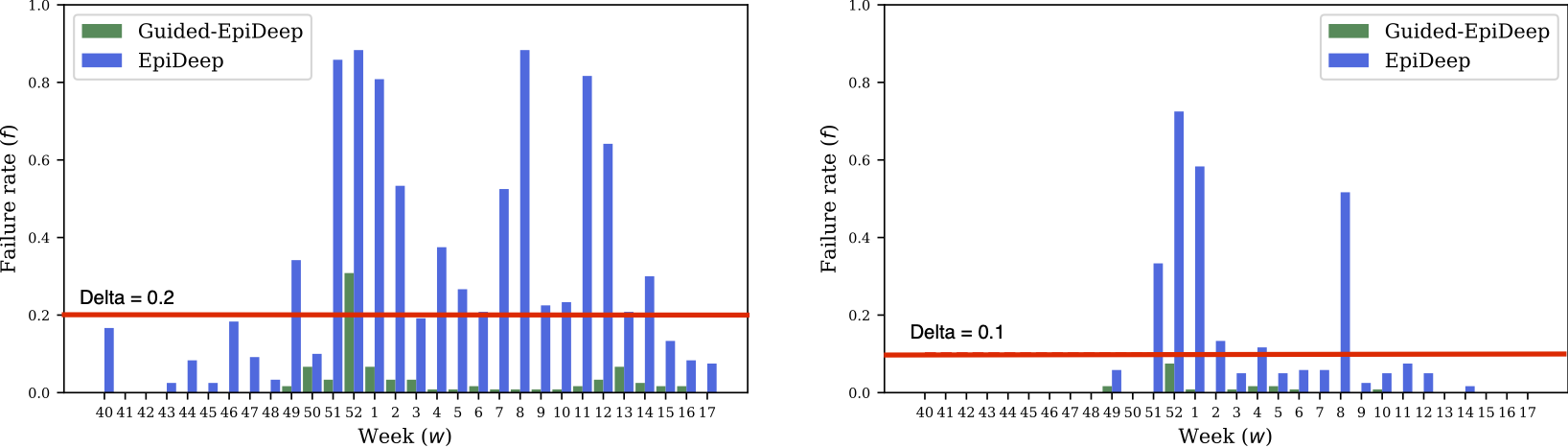}
    \caption{Performance of \alg{} in specific guidance. Figures show failure rate ($\f$) for different combinations of $\epsilon$ and $\delta$: (left) $\epsilon=0.25$ and $\delta=0.2$; (right) $\epsilon=0.5$ and $\delta=0.1$.
    Guided EpiDeep is successful incorporating expert guidance in epidemic task $\mathcal{T}_w$ for every week $w$ in standard flu season as it is mostly within the bounds given by $\delta$.
    Note that $\f$ in EpiDeep is higher than the required tolerance $\delta$, but \alg{} is able to exhibit the desired behavior within the required tolerance.}
    \label{fig:epsilon_per_week}
\end{figure*}

Here, we want to quantify the rate at which \alg{} is able to ensure that the behavior imposed by the guidance is met in the test set. To do so, here we split the historical seasonal influenza data into the training set $D$ which consists 80\% of the seasons and test set $T$, which has the remaining seasons. \alg{} is trained on $D$ with the smoothness constraint with a $\epsilon = 0.25$ and $\delta = 0.2$ to return a model $\theta$. We repeat the experiment to make forecasts starting at week 40 of the epidemiological season till week 17. We then measure the failure rate, as defined earlier, of $\theta$ on the test set $T$ for each week. Then we repeat the experiment with $\epsilon = 0.5$ and $\delta = 0.1$. The result is presented in Figure~\ref{fig:epsilon_per_week}.

As seen in both Figure~\ref{fig:epsilon_per_week}, for both settings, \alg{} almost always ensures that the behavior imposed by the guidance is carried to the test data $T$. We observe that only 1 out of 80 observations, only one \alg{} has a failure rate higher than the threshold $\delta$. On the other hand, the baseline EpiDeep has significantly higher failure rate consistently, with the failure rate for many observations greater than $\delta$. This experiment demonstrates that \alg{} ensures that the desirable behavior is observed while forecasting on test data, while the baselines fail to do so.

\par\noindent\textbf{Failure in Incorporation of Guidance.} In the rare case when the \alg{} fails to return a model (NSF) or the returned model does not ensure that the desirable behavior is observed in test data, as in week 52 in Figure~\ref{fig:epsilon_per_week} (left), the user is free to adjust one or more of the three knows our framework, data, model, and tolerance to allow the framework to search for a better forecasting model. For example, in the same example, setting a higher $\delta$ may ensure that the selected model satisfies the constraint in the test data as well.

\subsubsection{Case study: Holiday Correctness }

As mentioned earlier, during the holiday season in late December/early January the influenza incidence temporarily drops. Typically, the observed drop in the influenza incidence around the holiday season is not drastic. Hence, here we expect enforcing smoothness leads to better performance.

\begin{figure}[h]
\centering
  \begin{tabular}{cc}
    \includegraphics[width=0.25\textwidth]{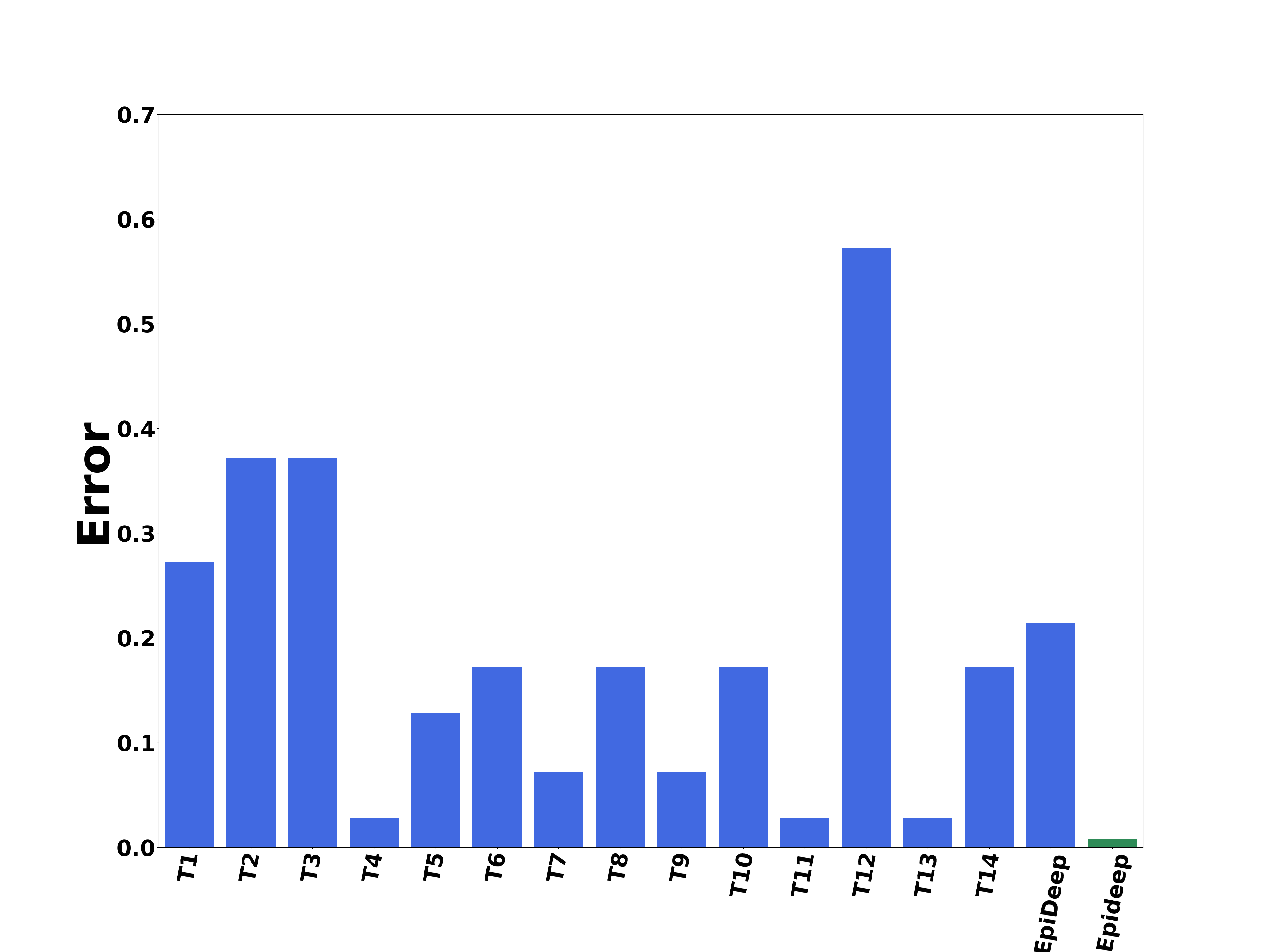} &
    \includegraphics[width=0.25\textwidth]{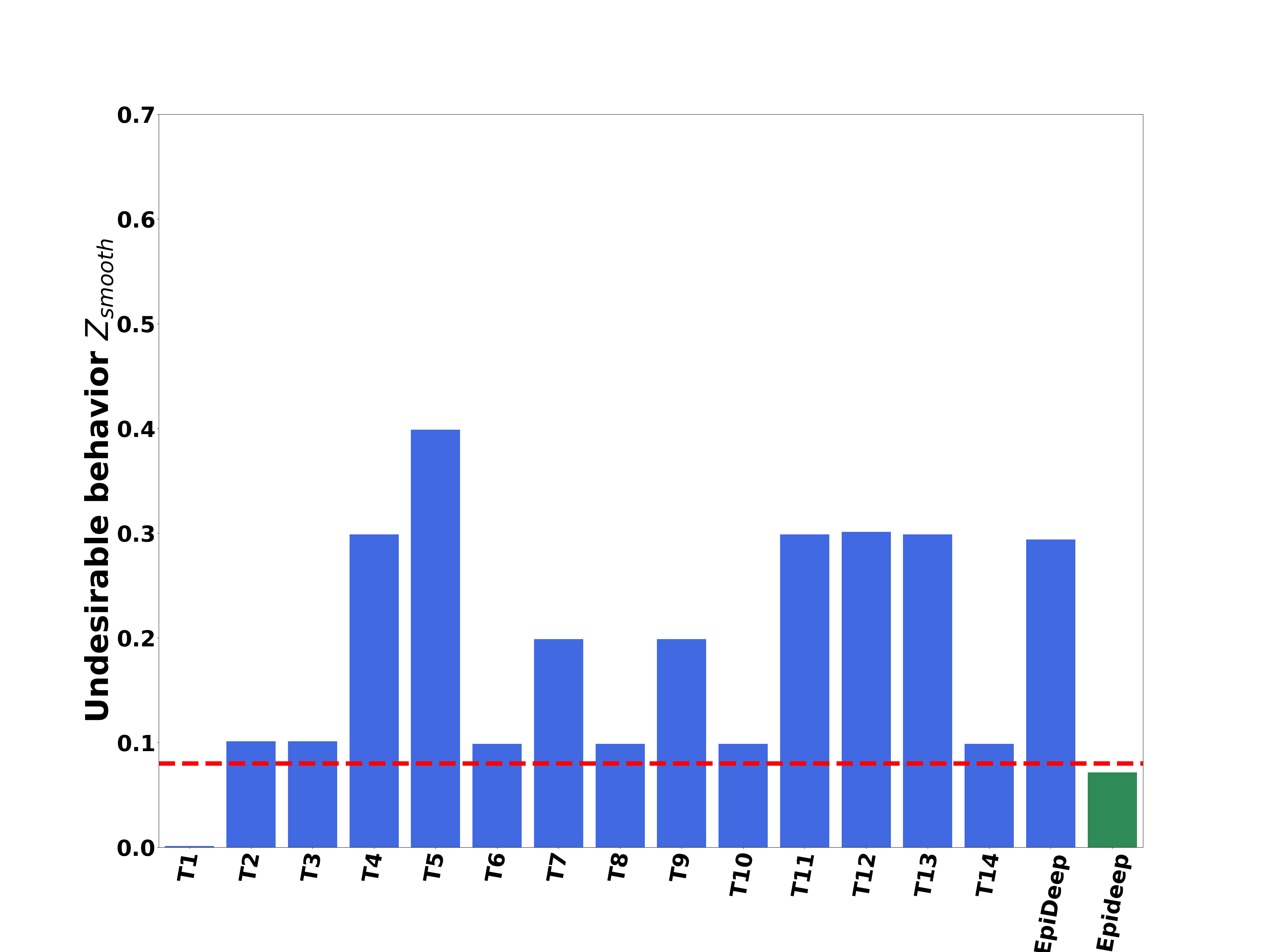} \\
    (a)  Performance  & (b) Guidance \\
  \end{tabular}
  
    \caption{Comparison of approaches in terms of (a) RMSE in forecasting and (b) $\zs$ for the smoothness constraint. n both plots lower is better. The red line in (b) is the threshold determined by guidance. T1 to T14 is the performance of teams participating in the 2015 FluSight challenge in HHS region 1.
    Guided-Epideep (GEpideep in the plot) is the only method to satisfy the smoothness constraint and maintain a low performance error.}
    \label{fig:case}
\end{figure}

Here we gathered the submission made by different models forecasting the influenza incidence in the first week of January 2016 from the 2015-2016 influenza season in the US national region as well as HHS Region 1. For each of the model, we measure the smoothness of prediction as the difference between the observed incidence in the last week of December and the forecast for the first week of January. Similarly, we also measure the performance of each model as the RMSE between the forecast and eventually observed ground truth incidence in the first week of January. We simulate the forecast for EpiDeep and our approach Guided-Epideep and compute both the metric. The result for the national region is presented in Figure \ref{fig:epsilon_intro} and the result for the HHS Region 1 is presented in Figure~\ref{fig:case}.

As observed in both Figures ~\ref{fig:epsilon_intro} and ~\ref{fig:case}, only Guided-Epideep and a baseline model $T4$ satisfy the smoothness constraint. All other methods fail to satisfy the constraint. Note that while $T4$ satisfies the constraint, it has a very high performance error due to the fact that it forecasts the incidence to be nearly the same as the last observed incidence. However, Guided-EpiDeep avoids this issue as ensures that the smoothness constraint is met, while the performance is also maintained. 

\subsection{Automatic Guidance}
\label{subsec:auto}

As mentioned earlier, in the automatic guidance mode, the user/expert provides the guidance. However, the other parameters are not known. Here, \alg{} searches for the ideal parameter set which can incorporate the guidance. Here, for automatic guidance, we evaluate \alg{} in terms of all the questions.

\subsubsection{Performance}
\begin{figure}[t]
    \centering
    \includegraphics[width=0.5\textwidth]{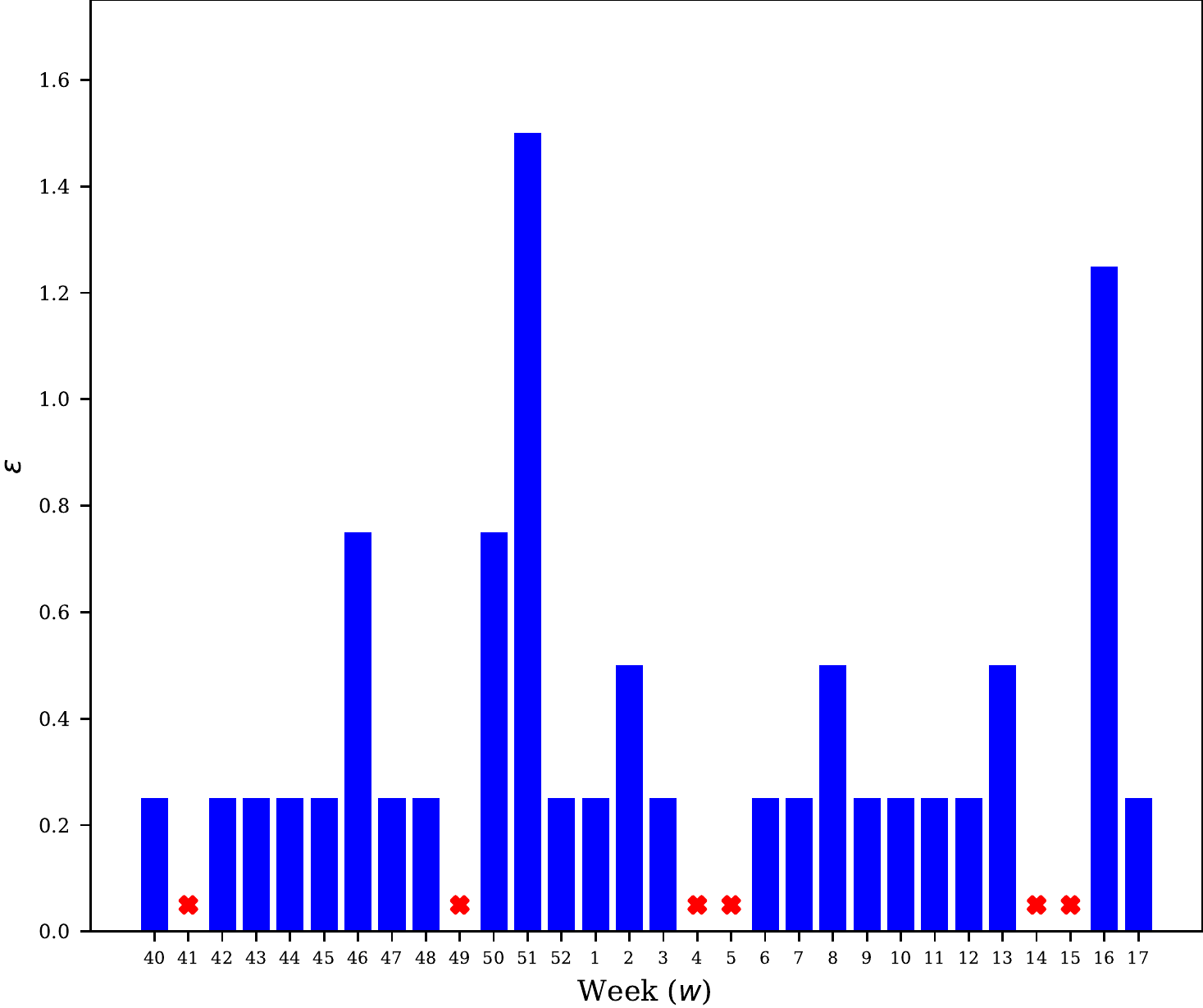}
    \caption{Automatic guidance over weeks. The y axis shows the value of $\epsilon$ found by \alg{} in automatic guidance mode. The red crosses represent the weeks where no suitable $\epsilon$ was found.}
    \label{fig:epsilonvRMSE}
\end{figure}

Here, we want to measure if \alg{} can find a parameter which can satisfy the constraints imposed by the guidance provided by the user. Here we use the same setup as in Direct guidance. We split the data into training $D$ and test $T$ sets with 4:1 ratio. The expert's requirement here is to ensure that the performance of \alg{} is better than the baseline model EpiDeep. We do so by enforcing that the ratio of RMSE of \alg{} to the RMSE of Epideep is less than 1. And the parameter to explore/detect is $\epsilon$.We repeated the experiment for each week in the influenza season. Figure \ref{fig:epsilonvRMSE} shows the result.

As seen in the figure, for most of the week \alg{} is able to find an $\epsilon$ such that the constraint defined by the user is met. Among, 40 weeks \alg{} fails to find $\epsilon$ in only 6 weeks, demonstrating that our framework is able to incorporate expert's guidance in the automatic guidance mode. For the weeks where $\epsilon$ could not be found, \alg{} communicates its inability to find a solution to the user.

\subsubsection{Case study: Regional Equity}

It has been well documented that forecasting models have better performance for some regions than others \cite{chakraborty_forecasting_2014}. Typically, regional error inconsistency is exhibited when comparing rural and urban regions. Therefore, an expert may want to make forecasting error balanced and achieve a more fair distribution of resources.  

In this case study, from the pool of models participating in the CDC Flusight challenge in 2018/19 season, the expert selects EpiDeep as the model for forecasting the current season. In the middle of the forecasting season, she realizes that EpiDeep performs well for HHS Region 1 but poorly for HHS Region 6. As she does not know the internals of EpiDeep, she prefers to use the automatic guidance mode, which allows her to figure out if regional equity can be incorporated without compromising shared accuracy.

\begin{figure}[t]
    \centering
    \includegraphics[width=0.5\textwidth]{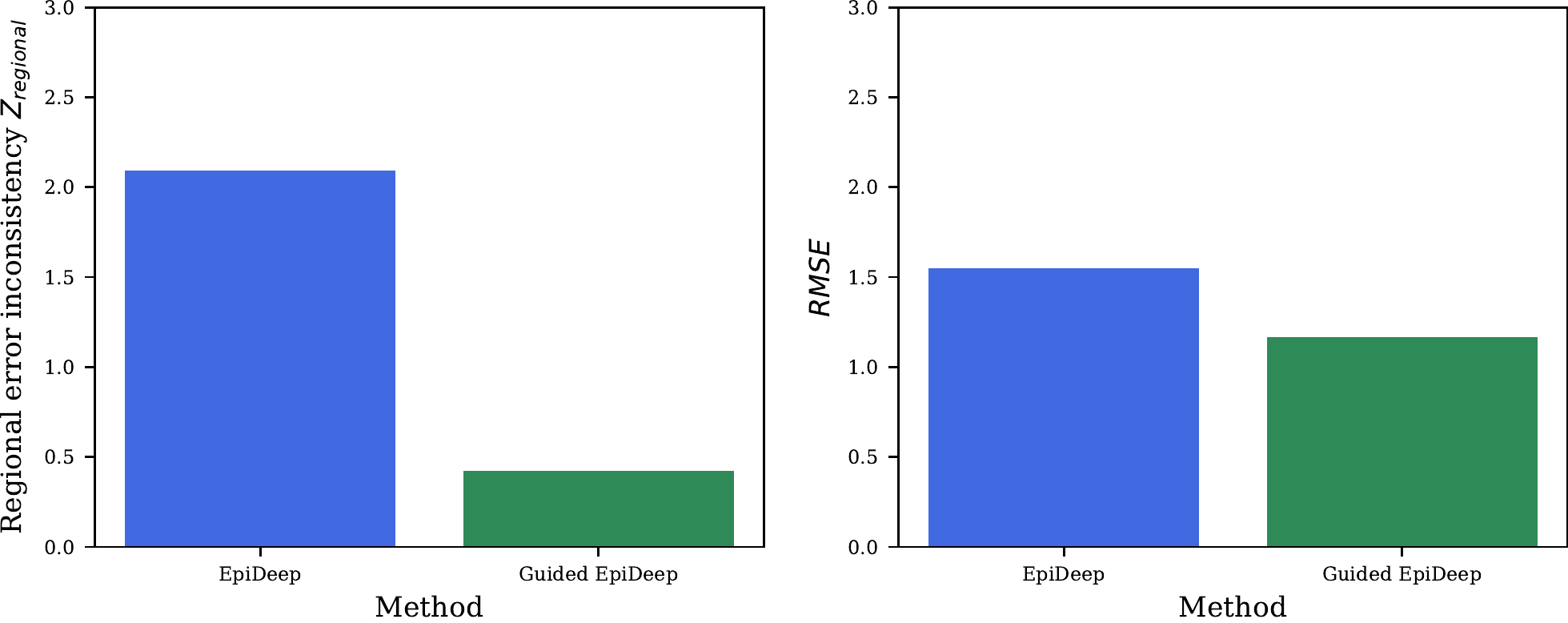}
    \caption{Automatic guidance for our case study on regional equity. Expert wants to make EpiDeep's predictions in Region 1 and Region 6 more equate/fair. Guided EpiDeep in automatic guidance finds out that for $\epsilon=1.0$, we are able to reduce regional inconsistency in addition to reduce average RMSE of the two models.}
    \label{fig:regional}
\end{figure}

To do so, here we run \alg{} with the regional equity consistency as discussed earlier for week $w=5$ in the automatic guidance mode. \alg{} returned $\epsilon=1.0$ and the model which performs best on train data in terms of accuracy. We then compare the performance of 
this model with Epideep both in terms of RMSE and regional equity consistency $\zr$. This result is summarized in Figure ~\ref{fig:regional}.

As shown in the figure, \alg{} is able to find an $\epsilon$ which is able to enforce the regional equity constraint. Moreover, it is also able to minimize RMSE. On the other hand, the baseline EpiDeep has high variance in the error in prediction between the regions and also has higher overall error. Hence, the automatic guidance mode of \alg{} is well suited for guidance incorporation when the parameters are unknown.

\section{Related Work}

\par \noindent \textbf{Epidemic Forecasting: } Epidemic forecasting models and generally categorized into statistical~\cite{ tizzoni2012real,adhikari2019epideep,osthus2019dynamic,brooks2018nonmechanistic} and mechanistic based approaches~\cite{shaman2012forecasting,zhang2017forecasting}. Ensemble of mechanistic and statistical approaches too have been proposed~\cite{reich2019accuracy}. There also has been interest in leveraging external data source in epidemic forecasting such as social media~\cite{chen2016syndromic,lee2013real}, search  engine~\cite{ginsberg2009detecting,yuan2013monitoring},  environmental and weather~\cite{shaman2010absolute,tamerius2013environmental}, and a combination of heterogeneous data~\cite{chakraborty2014forecasting}. 

Recently, there has been surging interest in leveraging deep learning for influenza forecasting. Adhikari et al.~\cite{adhikari2019epideep} proposed EpiDeep which leverages deep architecture to exploit seasonal similarity for epidemic forecasting. Similarly, Wang et al. proposed DEFSI~\cite{wang2019defsi} which exploits intra and inter seasonal data for forecasting. Other approaches like ~\cite{venna2017novel,volkova2017forecasting} have limited use case (example, for military population) and/or require external data sources (example, twitter, weather). However, none of these approaches are able to incorporate expert guidance.

\par \noindent \textbf{Time Series Analysis: } A field related to o epidemic forecasting in data mining is Time Series Analysis. Several approaches have been proposed such as auto-regression, kalman-filters and groups/panels~\cite{box2015time, sapankevych2009time,jha2015clustering}. Several deep learning approaches have also been used for time series analysis~\cite{hochreiter1997long, connor1994recurrent}.

\par \noindent \textbf{Guided prediction framework: } The Seldonian optimization framework~\cite{thomas2019preventing} discussed earlier presents a general framework for expert guided prediction framework. Based on the Seldonian framework, 
\citet{metevier2019offline} proposed Robinhood, an algorithm for fairness in a bandit setting. Several other approaches have been proposed for specific fairness objectives as well~\cite{joseph2016fairness,joseph2018meritocratic}. However, to the best of our knowledge, we are the first to introduce a guidance-based machine learning approach for epidemic forecasting.

\section{Conclusions}
\label{sec:conclusions}

In this paper, we study the novel general problem of incorporating expert guidance to statistical epidemic forecasting methods, using influenza prediction as an example. 
Leveraging the Seldonian optimization framework, we showcase a flexible, adaptable framework which can ensure that the given guidance can be incorporated subject to some probabilistic tolerance, whilst also maintaining performance accuracy. Additionally, our method also gives valuable feedback to the expert, if the guidance can not be successfully incorporated, to promote interactions. 
Via two natural guidance scenarios (smoothness and regional consistency) we show on real CDC surveillance data, that our method bounds the probability of undesirable behavior while also reducing RMSE by 17\%. As future work, one can focus on extending this framework to more types of guidance, and also handling probabilistic predictions (as opposed to point predictions we considered here).

\section*{Acknowledgements}
This paper is based on work partially supported by the National Science Foundation (Expeditions CCF-1918770, CAREER IIS-1750407, RAPID IIS-2027862, Medium IIS-1955883, DGE-1545362 and IIS-1633363), ORNL and Georgia Tech.

\bibliographystyle{ACM-Reference-Format}


\section*{Appendix}

The mapping between the teams in Figure 1 and 4 are as follows:
\begin{itemize}
    \item T1 → 4sight
    \item T2 → ARETE
    \item T3 → CU1
    \item T4 → CU2
    \item T5 → Delphi-Archefilter
    \item T6 → Delphi-Epicast
    \item T7 → Delphi-Stat
    \item T8 → Hist-Avg
    \item T9 → ISU
    \item T10 → JL
    \item T11 → KBSI1
    \item T12 → NEU
    \item T13 → UMN
    \item T14 → UnwghtAvg
\end{itemize}

\end{document}